# Learning a Clinically-Relevant Concept Bottleneck for Lesion Detection in Breast Ultrasound

Arianna Bunnell[1,2][0009-0000-6253-8402], Yannik Glaser[2], Dustin Valdez[1][0000-0001-9006-2966], Thomas Wolfgruber[1][0000-0001-8770-1800], Aleen Altamirano[3], Carol Zamora González[3] Brenda Y. Hernandez[1][0000-0003-0132-4042], Peter Sadowski[2][0000-0002-7354-5461] and John A. Shepherd[1][0000-0003-2280-2541]

[1] University of Hawaiʻi Cancer Center, Honolulu HI 96813, USA
jshepherd@cc.hawaii.edu
[2] University of Hawaiʻi at Mānoa, Honolulu HI 96822, USA
[3] Instituto Radiodiagnóstico, Managua, Nicaragua

**Abstract.** Detecting and classifying lesions in breast ultrasound images is a promising application of artificial intelligence (AI) for reducing the burden of cancer in regions with limited access to mammography. Such AI systems are more likely to be useful in a clinical setting if their predictions can be explained to a radiologist. This work proposes an explainable AI model that provides interpretable predictions using a standard lexicon from the American College of Radiology's Breast Imaging and Reporting Data System (BI-RADS). The model is a deep neural network featuring a concept bottleneck layer in which known BI-RADS features are predicted before making a final cancer classification. This enables radiologists to easily review the predictions of the AI system and potentially fix errors in real time by modifying the concept predictions. In experiments, a model is developed on 8,854 images from 994 women with expert annotations and histological cancer labels. The model outperforms state-of-the-art lesion detection frameworks with 48.9 average precision on the held-out testing set, and for cancer classification, concept intervention is shown to increase performance from 0.876 to 0.885 area under the receiver operating characteristic curve. Training and evaluation code is available at https://github.com/hawaii-ai/bus-cbm.

## 1 Introduction

Artificial intelligence (AI) is a promising tool for detecting and classifying lesions in breast ultrasounds, rivaling the accuracy of radiologists [1]. However, the adoption of AI systems for reviewing medical images is hindered by the inability of radiologists to verify predictions. Explainable AI (XAI) systems that can explain why a lesion has a high probability of being cancerous will help medical professionals identify situations in which the AI should not be trusted and perhaps enable them to correct the AI's mistakes. AI-empowered solutions have the potential to speed up reading and improve workflow for resource-limited scenarios, where there may be a single radiologist serving a large population. XAI may further improve workflow efficiency, allowing the radiologist to review cases where AI recommends biopsy with explanation, designating the AI as a second reader [2].

Concept bottleneck models (CBM) [3] are a type of neural network architecture which seeks to enforce interpretability by aligning intermediate model representations with human-defined concepts. By forcing the model to learn an intermediate representation based on medically-relevant concepts, the model predictions become both interpretable and modifiable. Learned intermediate concepts in neural networks do not usually align well with human-understandable concepts [4, 5], but CBM models can be designed with only human-understandable concepts in an intermediate bottleneck layer. Here, we explore both strict CBMs and models that can also make use of a side channel; a single node in the bottleneck layer which is not associated with any concept and only learned based on final classification.

The BI-RADS masses lexicon for ultrasound has five properties that characterize lesions: shape, orientation, margin, echo pattern, and posterior features. Each property is divided into sub-categories which are assigned to lesions to determine risk of malignancy and describe lesion characteristics. This language is familiar to radiologists and as such may be a useful way to communicate AI decisions and build trust in AI-based clinical decision support. The BI-RADS masses lexicon for ultrasound has been explored in breast ultrasound (BUS) image classification-only AI as a multi-task learning problem [6, 7]. While that approach provides interpretable predictions, the lack of bottleneck layer means that predictions cannot be updated by modifying the concepts. This modelling paradigm is most similar to the nonlinear CBM with a side channel we propose, however BI-RADS-NET [6, 7] fails to provide lesion localization or delineation.

"BI-RADS inspired" radiomic features have previously been used to predict lesion malignancy from known lesion delineations [8-12]. In these approaches, the BI-RADS mass features (as defined by the ACR) are approximated by computational methods. For example, posterior feature presence can be approximated by computing the difference in average gray pixel intensity between the lesion and its posterior area [11]. We propose a method which works from the BI-RADS features as defined by the ACR and does not require a priori lesion delineation or detection for malignancy prediction, improving both radiologist understanding and workflow efficiency.

Models using expertly-annotated BI-RADS features from known lesion boundaries to predict malignancy have been proposed [12-14] ([12] includes both morphometric and clinical BI-RADS features). We propose a method which does not require a priori lesion delineation or detection for malignancy prediction, improving workflow efficiency. This framework represents only the post-bottleneck architecture in the proposed method with no side channel, trained directly from expert annotations.

The main contributions of this paper are: (1) we propose a concept bottleneck approach to breast lesion classification from ultrasound using the BI-RADS masses lexicon; (2) we demonstrate the efficacy of this method on a dataset of 994 women; and (3) we release the first publicly-available AI model with mask-style outputs for lesion detection in BUS. All model predictions of malignancy are explicitly interpretable using a language familiar to radiologists and radiologists can easily update the AI predictions by modifying the concepts. This improves workflow (by reducing "translation time" between the radiologist and AI) and increases radiologist trust in AI decisions consistent with BI-RADS feature indications.

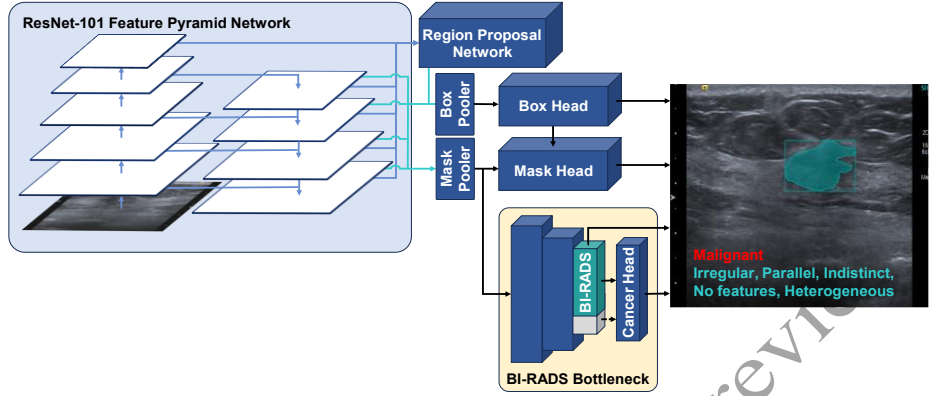

**Fig. 1.** An overview of BI-RADS CBM, including the Mask-RCNN underlying structure and the BI-RADS concept bottleneck sub-network.

## 2  BI-RADS Concept Bottleneck Network

We propose to integrate a CBM [3] into an established object detection architecture. Given a proposed lesion, our model (BI-RADS CBM) first predicts the BI-RADS masses lexicon, then uses it to predict whether the lesion is cancerous. For simplicity, we binarize the BI-RADS masses lexicon for each property into those classifications which are either indicative of malignancy or indicative of benignity. A lesion which has classifications indicative of benignity is oval shaped, oriented parallel to the skin, has a circumscribed (well-defined) margin, is anechoic, and has no posterior features (suggesting no difference in ultrasound wave speed through the lesion). All other classifications are binarized as being indicative of malignancy.

**Fig. 1** illustrates the architecture of the proposed BI-RADS CBM. In experiments, we train a complete AI model in three stages. We start from a standard Mask RCNN [15] architecture with a ResNet-101 feature pyramid network (FPN) [16, 17], pretrained on MS-COCO [18]. First (Stage 1), the model is fine-tuned to detect lesions. Second (Stage 2), a classification head is trained to predict the BI-RADS masses lexicon. Third (Stage 3), a cancer classification module is trained. All weights from previous stages are frozen during training. Models are implemented in PyTorch [19] using the Detectron2 [20] library.

The first stage (pre-bottleneck) of the cancer head is convolutional, with max-pooling. The second stage (post-bottleneck) is fully-connected. We hypothesize that a convolutional architecture from the mask feature maps enables the cancer head to make use of local (by pooling) and global (by FPN) information most effectively. Global information may enhance cancer prediction, particularly in a CBM with a side channel, by encoding information about cancer risk in breast tissue, such as breast density, the strongest risk factor for breast cancer outside of age [21]. The cancer head is trained on the outputs from the bottleneck layer, rather than directly from radiologist annotations.

## 3     Experiments & Data

### 3.1    Dataset

BUS images were collected from a prospective repository of 123,000 participants undergoing breast imaging between 2009-2023 in the Hawai'i and Pacific Islands Mammography Registry (WCG IRB, study number 1264170). Women in this registry are matched to the Hawai'i Tumor Registry (HTR) to determine cancer status. Inclusion criteria for all participants were as follows: (1) female; (2) has a record of diagnostic or screening BUS imaging; (3) BUS imaging has ≥ 2 DICOM image records; (4) exam rated as BI-RADS 2 (benign) or higher. Cases and controls were randomly selected from the pool of eligible women and matched 1:3 on birth year and BUS machine manufacturer. Cases were subject to the following additional inclusion criteria: (1) Record of invasive breast cancer diagnosis in the HTR (2) breast cancer diagnosis within one year of BUS imaging date; (3) BUS DICOM laterality matches HTR tumor laterality or is unknown. Controls were considered for inclusion if they failed to link to the HTR for cancer of any type. A total of 261 cases and 783 matched controls were included.

BUS images were annotated with lesion delineation, imaging artifacts, and the ACR BI-RADS masses lexicon for ultrasound by a breast radiologist with eight years of clinical experience in BUS image interpretation (A.A.). The radiologist was blinded to histological cancer status and patient identifier. Images were annotated with adapted VIA Annotation Software [22]. A total of 10,291 BUS images were annotated. To determine the reliability of the radiologist annotations, we constructed a concurrence reading set to measure inter-radiologist variability. 900 images, balanced between histologically benign, normal (benign, with no annotations from A.A. and histologically malignant, were sampled and read by C.Z. Reads were performed under the same blinding protocol as A.A. Inter-observer agreement was measured with Cohen's κ [23]. For lesion existence, inter-observer agreement was found to be 0.784 (substantial agreement).

To measure binarized BI-RADS inter-observer agreement, only annotations where both radiologists a) delineated a lesion and b) those delineations overlap with intersection over union (IoU) 0.25 are considered. A total of 620 lesions were considered in this calculation (A.A. and C.Z. delineated a total of 705 and 741 lesions, respectively). Cohen's κ values were 0.675 (substantial agreement), 0.568 (moderate), 0.730 (substantial), 0.701 (substantial), and 0.309 (fair) for binarized lesion shape, orientation, margin, echo pattern, and posterior features, respectively. The network is trained with annotations from one expert (A.A.).

### 3.2    Data Preprocessing

After annotation, BUS images which were identified as containing clips/markers (n = 43), biopsy needles (n = 377), and breast implants (n = 109) were excluded. Additionally, images whose DICOM header indicated that they were collected during breast biopsy (n = 59), invalid images (little/no breast tissue visible, n = 294), images collected with elastography (n = 203), images missing cancer status linkage (n = 9), images with incomplete annotations (n = 171), and images where the radiologist was unsure about lesion location or any of the BI-RADS masses lexicon classifications for any lesion in the image (n = 172) were excluded. After all image-level exclusions, 163 complete and 98 incomplete case-control groups, containing 249 cases and 745 controls, remain.

Data were randomly split into training (70%), validation (10%), and testing (20%) by case-control group. No women, images, or lesions are represented in more than one split. **Table 1** displays a summary of women, lesions, and images per data split and overall. The supplement contains additional summary statistics for the study sample, including BI-RADS masses lexicon distribution within each split. Note that lesion counts represent the number of total annotated areas, not the number of unique lesions.

**Table 1.** Characteristics of the study sample. Additional sub-population counts and characteristics can be found in the supplement. *All image-level counts reflect the number of images collected, not the resulting number of images after dual-view images were split. Dx = diagnosis.

| Dataset Characteristic, Unit | Train | Validation | Test | Overall |
|---|---|---|---|---|
| Women with benign findings, N (%) | 520 (75.0) | 75 (74.3) | 150 (75.0) | 745 (74.9) |
| Women with malig. findings, N (%) | 173 (25.0) | 26 (25.7) | 50 (25.0) | 249 (25.1) |
| Median days btw. BUS & Dx (IQR) | 4.0 (24.0) | 2.5 (22.0) | 0 (24.8) | 3.0 (24.0) |
| Mean no. of images/woman, N (SD) | 9.03 (4.89) | 9.01 (3.87) | 8.42 (3.33) | 8.91 (4.52) |
| Images, N* | 6,260 | 910 | 1,684 | 8,854 |
| Images with two views, N (%) | 830 (13.3) | 112 (12.3) | 119 (7.1) | 1,061 (12.0) |
| Mean no. of lesions/image, N (SD) | 1.26 (0.50) | 1.21 (0.41) | 1.17 (0.40) | 1.24 (0.48) |
| Mean no. of views/woman, N (SD) | 8.39 (6.89) | 7.54 (5.06) | 6.51 (3.92) | 7.94 (6.29) |
| Lesions with benign findings, N (%) | 2,626 (62.5) | 369 (64.4) | 584 (67.0) | 3,579 (63.4) |
| Lesions with malig. findings, N (%) | 1,577 (37.5) | 204 (35.6) | 288 (33.0) | 2,069 (36.6) |

### 3.3 Experiments

All experimental model specifications are trained with image augmentations (random brightness, random horizontal and vertical flipping, random contrast, and random cropping). Training minimized the binary cross-entropy loss for cancer and per-concept concept classification using the SGD optimizer with mini-batches of 16 images per batch and a linear learning rate warmup (0.001). Complete model configuration files can be found in the released code repository. Each training stage is undertaken with the weights from the previous stage frozen.

**Cancer head complexity.** The architecture of the cancer prediction head in the BI-RADS CBM is varied to examine the tradeoff between model explainability and performance in cancer prediction. The three architectures are: (1) a linear cancer head from the concepts only, representing the most direct clinically-interpretable prediction; (2) a nonlinear cancer prediction head from the concepts only; and (3) a nonlinear cancer prediction head wherein an additional, learnable, non-clinical concept is included in the bottleneck. The last architecture represents the least clinically-interpretable prediction.

**Corrected concepts.** Concepts are considered correctly predicted by the bottleneck if their intermediate representation (binarized at 0.5) corresponds to the correct class. When these intermediate activations are logits, they are sigmoid transformed for adjustment, then transformed back to logit space. When concepts are incorrectly predicted, they can be intervened on. We define two methods of intervention: minimal and maximal. In minimal and maximal correction, representations are adjusted just until the correct class is predicted with pseudo-probability 0.51 and 0.99, respectively.

**Comparison with baseline model.** We train an additional, non-explainable cancer head on top of the backbone as a baseline. We do not evaluate any of the publicly available frameworks for lesion detection in BUS on our data. CVA-Net [24] and [25] are not well-suited for our data due to our lack of temporally-ordered BUS frames. [26] and [27] fail to provide segmentation mask-style lesion detections, limiting direct comparability to BI-RADS CBM. We provide comparisons to reported results only.

### 3.4    Performance Evaluation

Performance is evaluated per stage of training for the concept bottleneck model. For the lesion detection task (Stage 1), performance is evaluated using standard object detection metrics for both segmentation mask and bounding box targets. We report average precision (AP), $AP_{50}$, and $AP_{75}$ for each target. The maximum number of detections is set at 10 for all evaluations. For the concept (Stage 2) and cancer (Stage 3) classification tasks, performance is evaluated via area under the receiver operating characteristic curve (AUROC) with 95% confidence intervals, computed using DeLong's method [28, 29], at IoU 0.5 and 0.75.

**Evaluation with corrected concepts.** To assess the potential clinical utility of intervenable concepts, we compute cancer classification performance (via AUROC) with corrected concepts. When one ground truth annotation exists in an image, all lesions' incorrectly predicted concepts are adjusted to the ground truth values, according to the adjustment type (minimal or maximal). When more than one ground-truth annotation exists in an image, each detected lesion is matched to a ground-truth annotation based on maximum IoU. If IoU is 0, the predictions are not changed. When no ground-truth annotations exist in the image, none of the predicted concepts are adjusted.

## 4    Results

Model hyperparameters were systematically searched using Optuna [30], with 25 trials undertaken for each training stage. The final model was chosen based on performance on the validation set. Hyperparameter search space and results are reported in the

Learning a Clinically-Relevant Concept Bottleneck for Lesion Detection in Breast Ultrasound 7

Supplement. The results from the lesion detection task are reported in **Table 2**. The BI-RADS CBM backbone detected lesions with AP 0.469 for bounding box-style detections on the held-out testing set, outperforming both video-based baselines on their respective testing sets. Notably, the BI-RADS CBM backbone also significantly outperforms the reported image-level results reported in [25]. Performance characteristics for BI-RADS clinical concept prediction are reported in **Table 3.**

Cancer classification results for the concept correction and cancer head complexity experiments are reported in **Table 4**. As expected, the best-performing model without intervention was the most flexible (and least interpretable) non-linear model with a side-channel (0.875 AUROC at IoU 0.5). Removing the side channel slightly reduces performance but increases interpretability of the model. Without the side-channel, it is possible to perform counterfactual reasoning by modifying individual concepts and observing the output prediction, with the non-linear version having slightly higher performance (0.865 AUROC at IoU 0.5) than the linear version (0.863 AUROC at IoU 0.5) at the cost of the relationship between concepts and predictions being less interpretable. Notably, all BI-RADS CBM model designs outperformed the baseline model which did not use BI-RADS concepts at all (0.850 AUROC at IoU 0.5). The BI-RADS features add domain knowledge that encourages the model to capture visual features indicative of cancer status, rather than learning all features from scratch.

**Table 2.** Performance characteristics for the lesion detection task. Segm = segmentation mask performance metrics. BBox = bounding box performance metrics. Performance metrics are reported as in original papers for each method. [27] and [26] do not report AP and so are excluded from this table.

| Model/Framework | AP | | $AP_{50}$ | | $AP_{75}$ | |
|---|---|---|---|---|---|---|
| | Segm | BBox | Segm | BBox | Segm | BBox |
| BI-RADS CBM | 0.489 | 0.469 | 0.780 | 0.775 | 0.554 | 0.528 |
| STNet [25] | N/A | 0.400 | N/A | 0.703 | N/A | 0.433 |
| CVA-Net [24] | N/A | 0.361 | N/A | 0.651 | N/A | 0.385 |

**Table 3.** Performance characteristics for the concept classification task (95% CI shown in parentheses).

| BI-RADS Mass Lexicon Concept | AUROC @ IoU = 0.50 (n = 807) | AUROC @ IoU = 0.75 (n = 616) |
|---|---|---|
| Lesion Posterior Features | 0.616 (0.572, 0.659) | 0.551 (0.501, 0.601) |
| Lesion Echo Pattern | 0.921 (0.903, 0.939) | 0.928 (0.908, 0.948) |
| Lesion Shape | 0.901 (0.876, 0.927) | 0.897 (0.864, 0.930) |
| Lesion Orientation | 0.842 (0.798, 0.887) | 0.838 (0.777, 0.898) |
| Lesion Margin | 0.916 (0.893, 0.940) | 0.915 (0.885, 0.945) |

The effect of concept intervention was mixed. The maximal concept correction strategy unilaterally degraded model performance in cancer classification. One likely reason for this is that the cancer classifier was trained on soft labels from the CBM, and so the model is forced to generalize beyond its training data distribution in the maximal intervention strategy. In contrast, the minimal strategy always increased model performance, but more for the linear and non-side channel cancer heads. This suggests that the side channel cancer head ignores the predicted BI-RADS features in favor of the learned side channel.

**Table 4.** Performance characteristics for the cancer classification task, with and without concept correction on the testing set. CBM = BI-RADS CBM.

| Model | Side channel? | Non-linear? | Correction? | AUROC @ IoU = 0.50 | AUROC @ IoU = 0.75 |
|---|---|---|---|---|---|
| CBM | No | No | None | 0.863 (0.833, 0.892) | 0.861 (0.824, 0.898) |
| CBM | No | No | Minimal | **0.885 (0.857, 0.912)** | **0.885 (0.851, 0.919)** |
| CBM | No | No | Maximal | 0.839 (0.808, 0.871) | 0.841 (0.802, 0.879) |
| CBM | No | Yes | None | 0.865 (0.835, 0.894) | 0.862 (0.825, 0.899) |
| CBM | No | Yes | Minimal | 0.875 (0.846, 0.905) | 0.874 (0.836, 0.912) |
| CBM | No | Yes | Maximal | 0.823 (0.789, 0.857) | 0.814 (0.770, 0.858) |
| CBM | Yes | Yes | None | 0.875 (0.847, 0.903) | 0.871 (0.836, 0.906) |
| CBM | Yes | Yes | Minimal | 0.875 (0.847, 0.903) | 0.872 (0.837, 0.907) |
| CBM | Yes | Yes | Maximal | 0.851 (0.819, 0.882) | 0.845 (0.806, 0.885) |
| **Baseline** | N/A | N/A | N/A | 0.850 (0.821, 0.879) | 0.876 (0.845, 0.906) |

## 5 Conclusion

To enhance interpretability of cancer status classification of lesions in BUS, while simultaneously providing automatic lesion delineation, we propose BI-RADS CBM. Experiments on a large internal dataset of BUS imaging with expert annotations and histological cancer labels demonstrate that an AI model can be both accurate and interpretable. We further demonstrate for the first time that BI-RADS concept intervention is possible and increases cancer classification performance. BI-RADS CBM presents an XAI solution for lesion detection, description, and classification from BUS suitable for use in limited-resource scenarios to stretch limited radiological resources.

**Acknowledgments.** This study was funded by the NCI (grant number 5R01CA263491).

**Disclosure of Interests.** The authors have no competing interests to declare that are relevant to the content of this article.

# Supplemental Information for Learning a Clinically-Relevant Concept Bottleneck for Lesion Detection in Breast Ultrasound

**Table S1.** Additional characteristics of the study sample. *All image-level counts reflect the number of images collected, not the resulting number of images after dual-view images were split. BUS = breast ultrasound. Dx = diagnosis. BI-RADS feature lesion counts are reported as dichotomized for model training and evaluation, not as defined by the ACR [S1].

| Dataset Characteristic, Unit | Train | Validation | Test |
|---|---|---|---|
| **Women, N** | 693 | 101 | 200 |
| Mean age at BUS, years (SD) | 63.6 (12.8) | 62.9 (11.7) | 63.3 (13.9) |
| Mean age at Dx, years (SD) | 63.7 (12.5) | 62.9 (12.2) | 63.0 (13.8) |
| **Images, N*** | 6,260 | 910 | 1,684 |
| Images with benign findings, N (%) | 4,587 (73.3) | 661 (72.6) | 1,307 (77.6) |
| Images with malig. findings, N (%) | 1,673 (26.7) | 249 (27.4) | 377 (22.4) |
| Images on PHILIPS system, N (%) | 2,881 (46.0) | 489 (53.7) | 1,113 (66.1) |
| Images on SIEMENS system, N (%) | 2,949 (47.1) | 381 (41.9) | 539 (32.0) |
| Images on ATL system, N (%) | 430 (6.9) | 40 (4.4) | 32 (1.9) |
| BI-RADS 1/2/3 images, N (%) | 3,914 (62.5) | 566 (62.2) | 1,159 (68.8) |
| BI-RADS 4 images, N (%) | 1,410 (22.5) | 235 (25.8) | 357 (21.2) |
| BI-RADS 5/6 images, N (%) | 638 (10.2) | 92 (10.1) | 137 (8.1) |
| BI-RADS 0/Unk. images, N (%) | 298 (4.8) | 17 (1.9) | 31 (1.8) |
| **Lesion Views, N** | 4,203 | 573 | 872 |
| Oval lesions, N (%) | 2,801 (66.6) | 368 (64.2) | 573 (65.7) |
| Irregular/round lesions, N (%) | 1,402 (33.4) | 205 (35.8) | 299 (34.3) |
| Parallel lesions, N (%) | 3,560 (84.7) | 449 (78.4) | 752 (86.2) |
| Not parallel lesions, N (%) | 643 (15.3) | 124 (21.6) | 120 (13.8) |
| Circumscribed lesions, N (%) | 2,937 (69.9) | 401 (70.0) | 598 (68.6) |
| Not circumscribed lesions, N (%) | 1,266 (30.1) | 172 (30.0) | 274 (31.4) |
| Anechoic lesions, N (%) | 1,259 (30.0) | 168 (29.3) | 339 (38.9) |
| Not anechoic lesions, N (%) | 2,944 (70.0) | 405 (70.7) | 533 (61.1) |
| Lesions w/o posterior feats., N (%) | 2,988 (71.1) | 383 (66.8) | 560 (64.2) |
| Lesions w/posterior feats., N (%) | 1,215 (28.9) | 190 (33.2) | 312 (35.8) |

**Table S2.** Hyperparameter search space and chosen hyperparameter values for all model training stages and variations. For each stage, the hyperparameters were tuned over 25 trials using Optuna's TPESampler [S2]. Frozen stage = stage at which the ResNet-101 FPN is frozen during training. FC = fully-connected. In Stage 2, the number of filters corresponds to the number in each convolutional layer in a residual block.

| Training Stage | Hyperparameter | Search Space | Value |
| --- | --- | --- | --- |
| **Concept Bottleneck Model** | frozen stage | {1, 2, 3, 4, 5}S | 4 |
| Stage 1: | # box head conv. layers | {1, 2, 3, 4, 5} | 3 |
| Lesion Detection | # box head FC layers | {1, 2, 3} | 4 |
|  | # mask head conv. layers | {1, 2, 3, 4, 5} | 5 |
|  | momentum | Uniform [0.1, 0.9] | 0.9 |
| Stage 2: | # filters in 1st layer | {512, 256, 128, 64} | 512 |
| Concept Classification | # filters in 2nd layer | {256, 128, 64, 32} | 64 |
|  | base learning rate | LogUniform [1e-7, 1e-1] | 0.093 |
|  | momentum | Uniform [0.1, 0.9] | 0.8 |
| Stage 3a: | FC layer width | {2048, 1024, 512, 256, 128, 64} | 512 |
| Cancer Classification | base learning rate | LogUniform [1e-7, 1e-1] | $4\times10^{-4}$ |
| (w/o side channel) | intermediate sigmoid | {True, False} | False |
|  | momentum | Uniform [0.1, 0.9] | 0.5 |
| Stage 3b: | base learning rate | LogUniform [1e-7, 1e-1] | $6\times10^{-6}$ |
| (w/ side channel) | momentum | Uniform [0.1, 0.9] | 0.8 |
| **Baseline Model** | base learning rate | LogUniform [1e-7, 1e-1] | $7\times10^{-2}$ |
|  | momentum | Uniform [0.1, 0.9] | 0.1 |

## Supplemental References